\documentclass[letterpaper, 10 pt, conference]{ieeeconf}
\usepackage[utf8]{inputenc}

\usepackage{amsmath, amssymb}
\usepackage{graphicx}
\usepackage{xcolor}          

\usepackage{float}

\usepackage[percent]{overpic}

\title{\LARGE \bf
A Hybrid Hinge–Beam Continuum Robot with Passive Safety Capping for Real-Time Fatigue Awareness
}

\author{Tongshun Chen$^{1}$, Zezhou Sun$^{1}$, Yanhan Sun$^{1}$, Yuhao Wang$^{1}$, Dezhen Song$^{1}$, Ke Wu$^{1}$ 
\thanks{$^{1}$Aurthors are with the Robotics Department, Mohamed bin Zayed University of Artificial Intelligence (MBZUAI), Abu Dhabi, United Arab Emirates {\tt\small ke.wu@mbzuai.ac.ae}}
}

\begin{document}

\maketitle
\thispagestyle{empty}
\pagestyle{empty}

\begin{abstract}

Cable-driven continuum robots offer high flexibility and lightweight design, making them well-suited for tasks in constrained and unstructured environments. However, prolonged use can induce mechanical fatigue from plastic deformation and material degradation, compromising performance and risking structural failure. In the state of the art, fatigue estimation of continuum robots remains underexplored, limiting long-term operation. To address this, we propose a fatigue-aware continuum robot with three key innovations: (1) a Hybrid Hinge-Beam structure where TwistBeam and BendBeam decouple torsion and bending: passive revolute joints in the BendBeam mitigate stress concentration, while TwistBeam's limited torsional deformation reduces BendBeam stress magnitude, enhancing durability; (2) a Passive Stopper that safely constrains motion via mechanical constraints and employs motor torque sensing to detect corresponding limit torque, ensuring safety and enabling data collection; and (3) a real-time fatigue-awareness method that estimates stiffness from motor torque at the limit pose, enabling online fatigue estimation without additional sensors. Experiments show that the proposed design reduces fatigue accumulation by about 49 
\% compared with a conventional design, while passive mechanical limiting combined with motor-side sensing allows accurate estimation of structural fatigue and damage. These results confirm the effectiveness of the proposed architecture for safe and reliable long-term operation.
\end{abstract}



\section{Introduction}

Cable-driven continuum robots (CDCRs) are lightweight and highly flexible manipulators with large workspaces, offering reliable and tractable displacement/force control through tendon-driven actuation with proximal placement of actuators.~\cite{russo2023overview,li2023review}. Compared with pneumatic~\cite{marchese2014soft}, hydraulic~\cite{zhang2022hydraulic}, and SMA-based actuation~\cite{xu2019sma}, cable-driven systems provide better stability and repeatability, and their actuation architecture naturally exposes motor-side force and position signals for state estimation and control~\cite{burgner2015continuum,rucker2011statics}. These features make CDCRs well-suited for minimally invasive interventions, inspection in confined spaces, and manipulation of delicate objects~\cite{burgner2015continuum,greigarn2018review,xu2021variable}.
\begin{figure}[t!]
    \centering
    \includegraphics[width=0.95\linewidth]{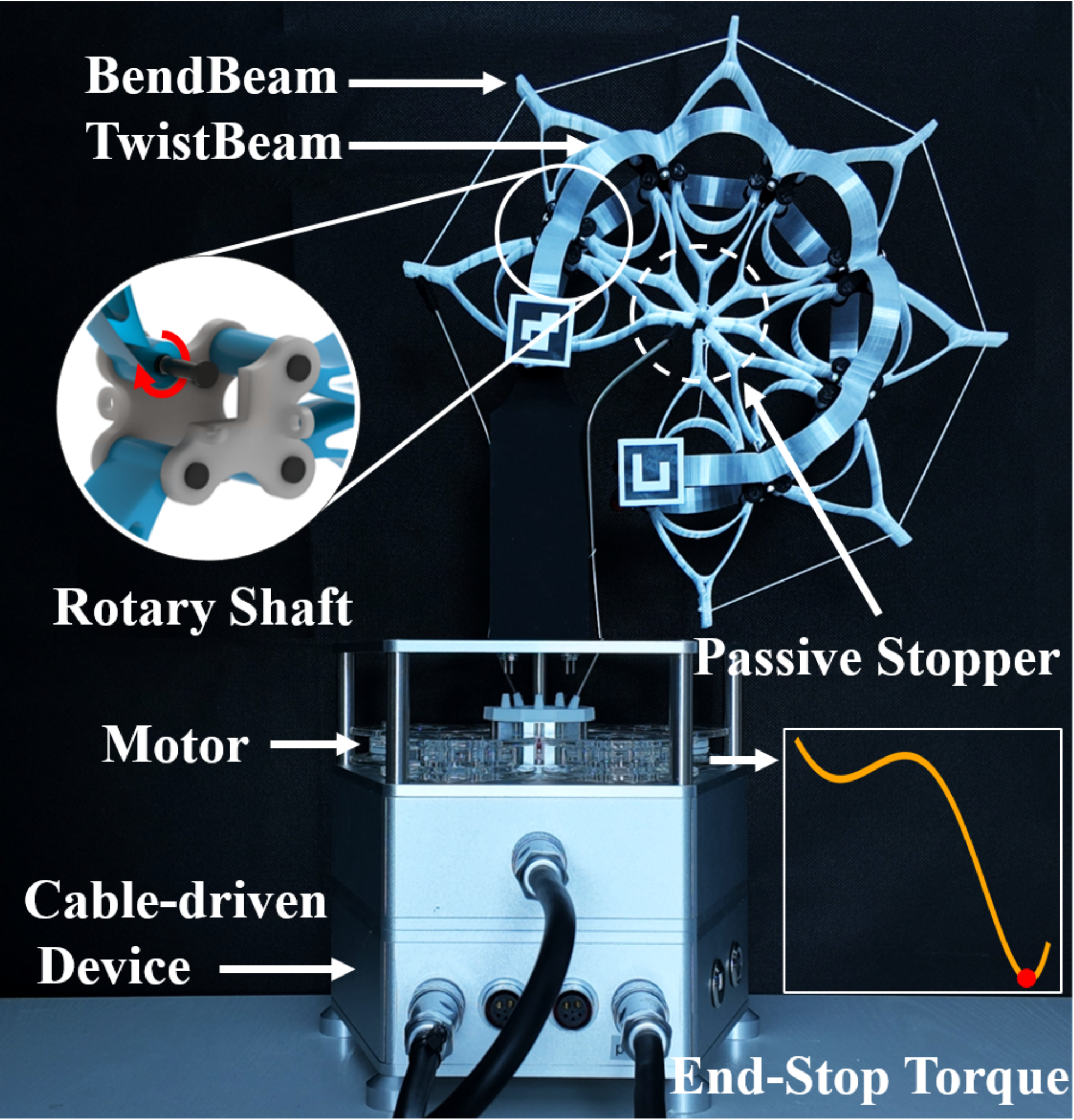}
    \caption{Overview of the proposed fatigue-aware CDCR}
    \label{fig:system_overview}
\end{figure}

To realize stable operation in the above mentioned applications, however, durability remains a major challenge. Long-term operation poses durability challenges: repeated loading and creep induce permanent plastic deformation and material degradation, reducing shape retention, load capacity, and motion accuracy~\cite{rucker2011statics, wang2021sma}.  In particular, fatigue estimation for continuum robots is still underexplored, and the absence of fatigue-aware design and monitoring strategies continues to limit their reliability in sustained operations~\cite{polly2020soft}.

\begin{figure*}[t]
    \centering
    \includegraphics[width=0.95\textwidth]{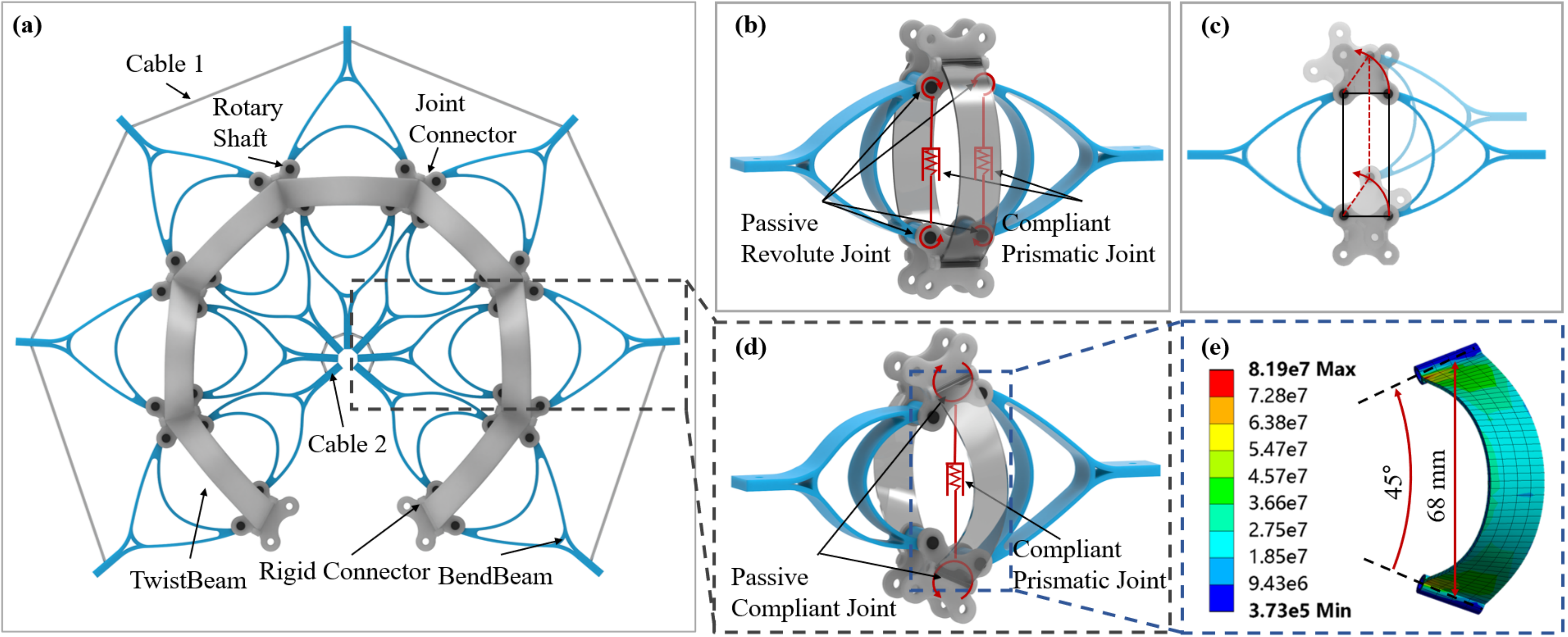}
    \caption{Hybrid hinge-beam architecture of the proposed fatigue-aware CDCR. 
    (a) Overall assembly with BendBeams, TwistBeams, and antagonistic cables; 
    (b) module detail highlighting the Passive Revolute Joint and Compliant Prismatic Joint; 
    (c) kinematic visualization of bending-induced offset; 
    (d) module detail with Passive Compliant Joint and Compliant Prismatic Joint; 
    (e) Finite Element Analysis (FEA) of the module under $45^\circ$ bending and $68\,\mathrm{mm}$ arc length.}
    \label{fig:hybrid_structure}
\end{figure*}


A major driver of fatigue in continuum structures is bending-induced stress concentration~\cite{howell2001compliant,trivedi2008soft}. To mitigate this, joint- and flexure-based designs have been proposed to redistribute strain~\cite{sun2022stress_distribution_ral}. For instance, \cite{clark2021endo,ma2024flexure} assembled continuum robots from discrete modular flexure joints, which help relieve stress concentration but introduce assembly errors and unstable performance. In contrast, \cite{dewi2024modular,sun2023topology,zhang2021compliant} increased the number of thin flexure hinges within each modular joint, thereby reducing the total number of joints needed to achieve similar stress-relief effects. However, these thin flexures are prone to local fracture due to insufficient strength, and the overall stiffness of the structure decreases. Alternatively, \cite{zhang2023hybrid,childs2021geometry} adopted monolithic designs that integrate repetitive or gradually varying flexure structures along the backbone, leading to more stable performance while reducing stress concentration, though still limited by low load capacity. Bio-inspired designs, such as knee-like structures, can distribute stress more uniformly, but complex intersecting flexure architectures increase fabrication difficulty, especially when extended to 3D implementations~\cite{sun2022stress}.   

Beyond improving structural durability, modeling or estimating fatigue in continuum robots is also critical. However, existing work remains rather limited. For example, \cite{wang2021sma} predicted failure trends of SMA-driven continuum robots using Weibull distribution based on experimental data, but such approaches are purely empirical and cannot enable real-time detection. it should be noted that cable-driven systems offer unique sensing advantages: actuator-side force and displacement signals are inherently available and can directly reflect the robot’s real-time state without embedded sensors~\cite{feliu2025actuation,zhenpu2025hybrid}. Nevertheless, to the best of our knowledge, the modeling of these signals for fatigue estimation in continuum robots has not yet been explored.

To address these challenges, we propose a fatigue-aware CDCR that combines a novel mechanical structure with a real-time fatigue estimation strategy (Fig.~\ref{fig:system_overview}). The proposed design enhances structural durability through stress redistribution and mechanical constraints, while the estimation method enables real-time fatigue awareness using only motor-side sensing. As shown in Fig.~\ref{fig:system_overview}, the main contributions of the paper are:  
\begin{itemize}[leftmargin=*, topsep=0pt, partopsep=0pt, parsep=0pt]  
    \item \textbf{Hybrid Hinge-Beam Design:} A structural design that decouples torsion and bending. BendBeams with passive revolute joints (Rotary Shaft) alleviate stress concentration, while TwistBeams redistribute loads, resulting in balanced deformation and enhanced durability.  
    \item \textbf{Passive Stopper:} A geometric-mechanical motion limiter combined with motor torque sensing. It ensures safe operation within actuation space and provides reliable feedback sensing data at the limit pose.  
    \item \textbf{Real-time Fatigue Awareness:} A strategy that estimates stiffness from torque at the stable limit pose (End-stop Torque), enabling real-time monitoring of fatigue progression and detection of structural damage.  
\end{itemize}  


\section{Mechanical Design} 

In this section, to improve durability and fatigue resistance, we propose a Hybrid Hinge-Beam design to optimize stress distribution, and a Passive Stopper mechanism that constrains the safe operational workspace while providing stable feedback data for continuum robots.


\subsection{Hybrid Hinge-Beam Design}

A Hybrid Hinge–Beam structure, composed of BendBeams and TwistBeams, is proposed to mitigate stress concentration and optimize stress distribution, thereby enhancing durability and fatigue resistance.


\subsubsection{BendBeam with Passive Revolute Joints}


Each BendBeam is connected to adjacent modules through passive revolute joints implemented with rotary shafts. A rigid connector holds the shafts coaxially, ensuring parallel alignment and defining the inter-connector spacing (Fig.~\ref{fig:hybrid_structure}). From a mechanical design perspective, this configuration enables large-angle bending while reducing stress concentration at the ends of the BendBeam~\cite{howell2001compliant}.



\begin{figure}[t]
    \centering
    \begin{overpic}[width=\linewidth]{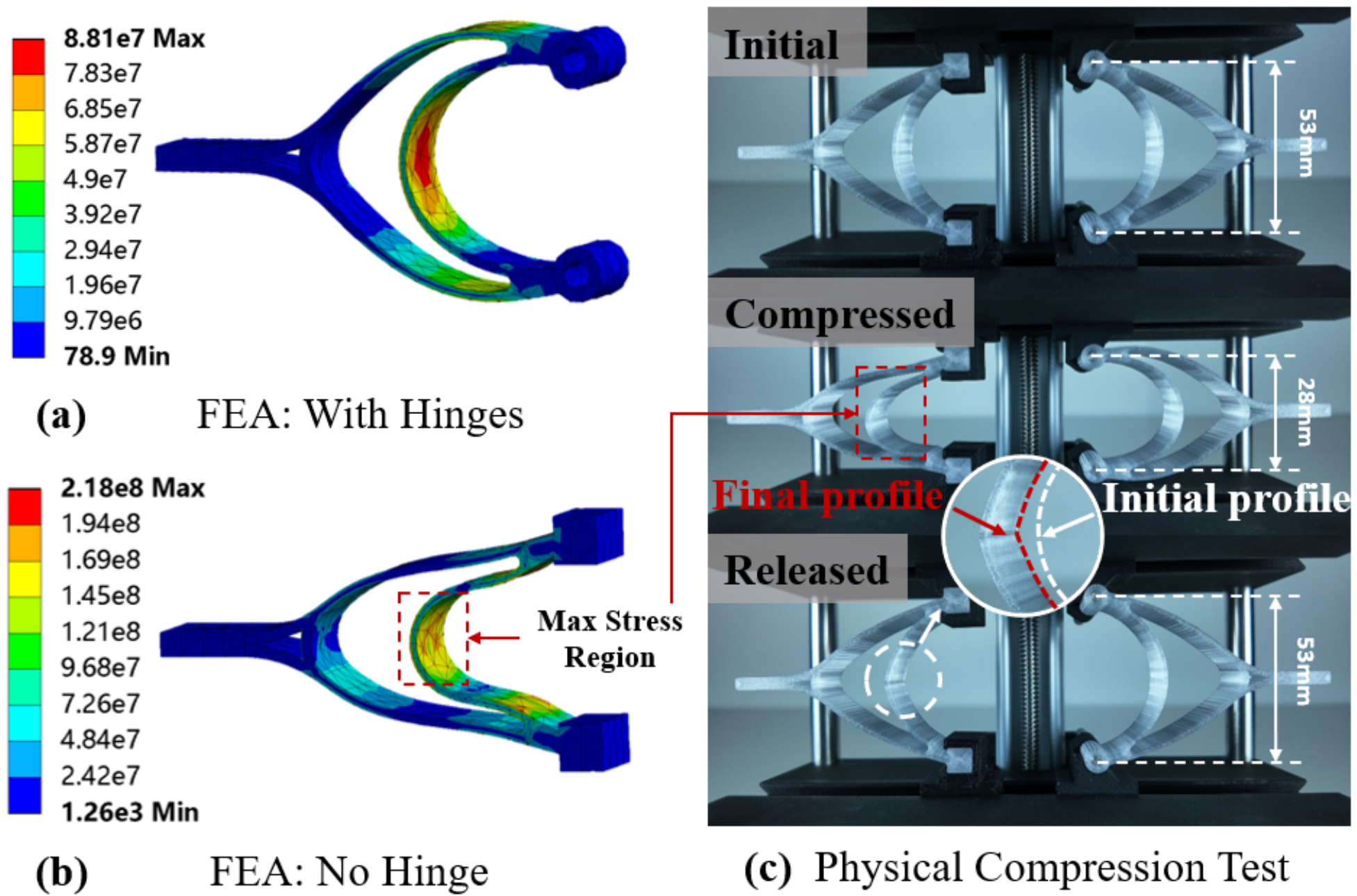}
    \end{overpic}
    \caption{
    FEA and physical testing of passive and non-passive revolute joint-integrated structures under 25\,mm of vertical compression.
    (a, b) Finite element models of both structures;  
    (c) Physical compression and recovery process. 
    }
    \label{fig:fig_rotational_hinge}
\end{figure}

\paragraph*{Finite Element Analysis and Experimental Validation} 
To validate this design concept, we performed both finite element analysis (FEA) of stress distribution and compression experiments to evaluate the elastic deformation range (Fig.~\ref{fig:fig_rotational_hinge}). The FEA results show that incorporating revolute joints significantly reduces the peak von Mises stress, from $2.18\times10^8$~Pa for the no-hinge beam to $8.81\times10^7$~Pa for the hinged beam under the same loading condition. The compression experiments further demonstrate that the no-hinge specimen exhibits irreversible curvature after $25\,\mathrm{mm}$ compression, whereas the hinged specimen fully recovers its shape, indicating a larger elastic deformation range and reduced susceptibility to fatigue caused by plastic deformation.



\paragraph*{Fatigue Tests}
To further evaluate fatigue resistance, BendBeam samples were cyclically compressed with amplitudes of $25$, $30$, and $35\,\mathrm{mm}$. Compression force and residual deformation were measured every 100 cycles. The results show that the $25\,\mathrm{mm}$ group maintained stable force and negligible residual deformation over $5{,}000$ cycles, the $30\,\mathrm{mm}$ group exhibited moderate degradation, and the $35\,\mathrm{mm}$ group failed after $2{,}500$ cycles. Therefore, $25\,\mathrm{mm}$ compression is identified as a safe and practical operating limit, which serves as a reference for subsequent design considerations.




\subsubsection{Role of the TwistBeam}  
While the BendBeam combined with revolute joints alleviates stress concentration, the revolute joints also introduce an additional lateral degree of freedom (DOF) according to the Grübler–Kutzbach criterion, which manifests as axial instability of the robot backbone (Fig.~\ref{fig:hybrid_structure}c). To suppress this undesired DOF and further optimize the stress distribution of the BendBeams, a TwistBeam is inserted between two BendBeams. As shown in Fig.~\ref{fig:hybrid_structure}d, the TwistBeam serves as two compliant constraints: (i) a compliant prismatic joint along its axis that shares axial and angular loading with the BendBeams, and (ii) a compliant torsional joint at the connector interface that stabilizes relative orientation and resists undesired angular rotation.

\begin{figure}[t]
\centering
\includegraphics[width=\linewidth]{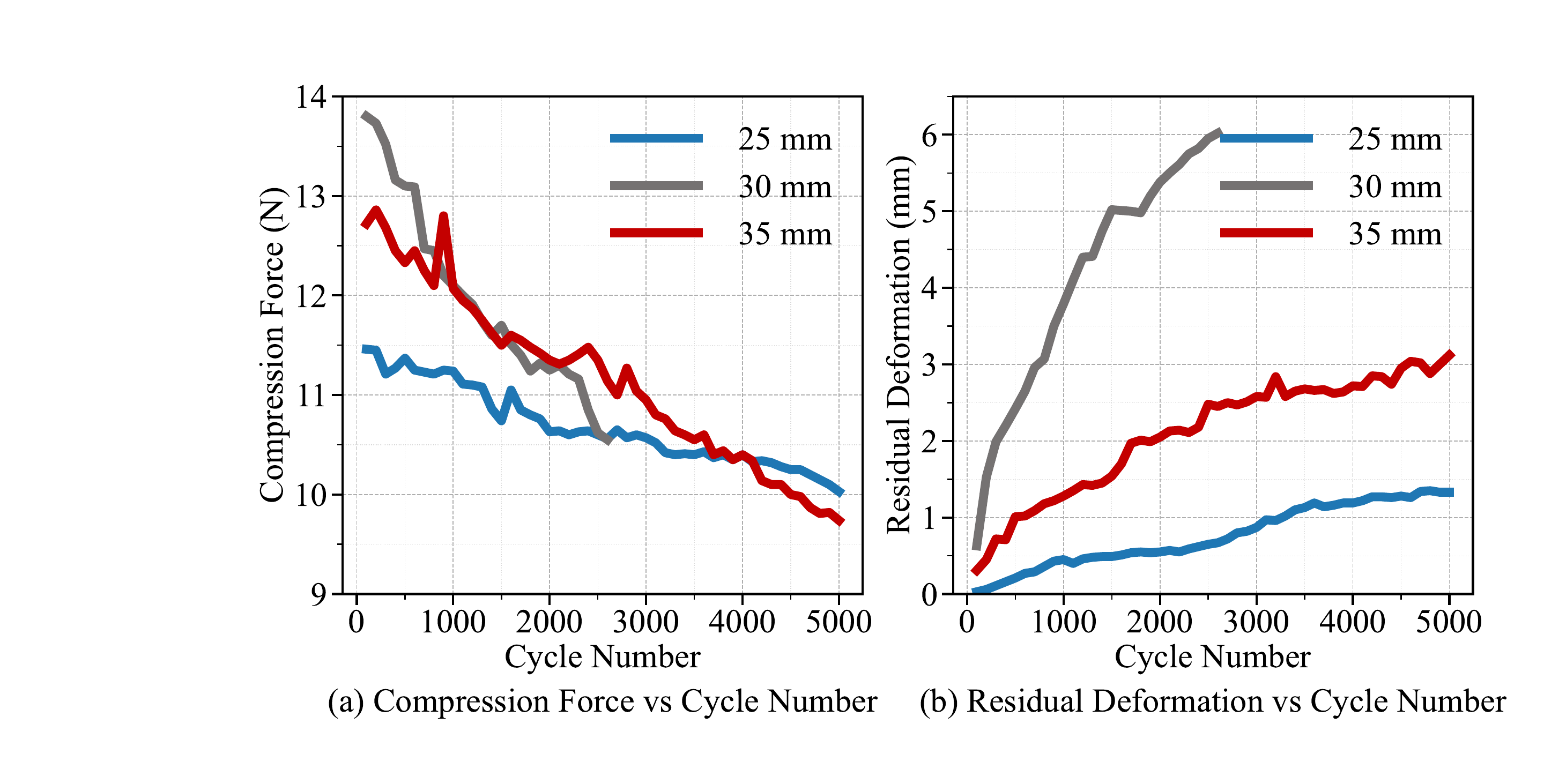}
\caption{
(a) Decay of force per cycle. 
(b) Accumulation of residual deformation.
Results are shown for displacement amplitudes of 25\,mm, 30\,mm, and 35\,mm.
}
\label{fig:fig_cycle}
\end{figure}

\paragraph*{Finite Element Validation}
FEA results under combined bending and axial compression confirm the safety margin of the design (Fig. \ref{fig:fig_cycle}). At $45^\circ$ bending with $4\,\mathrm{mm}$ axial shortening, the TwistBeam reaches a peak \textit{von Mises} stress of $8.19\times10^7$~Pa, which is lower than that of a BendBeam under $25\,\mathrm{mm}$ compression. Since BendBeams have been experimentally shown to withstand more than $5{,}000$ cycles at comparable stress levels, the TwistBeam is expected to remain fatigue-safe. Therefore, $45^\circ$ bending is defined as the design upper limit for safe deformation.

\subsection{Passive Stopper Design}

\begin{figure}[t]
    \centering
\includegraphics[width=\linewidth]{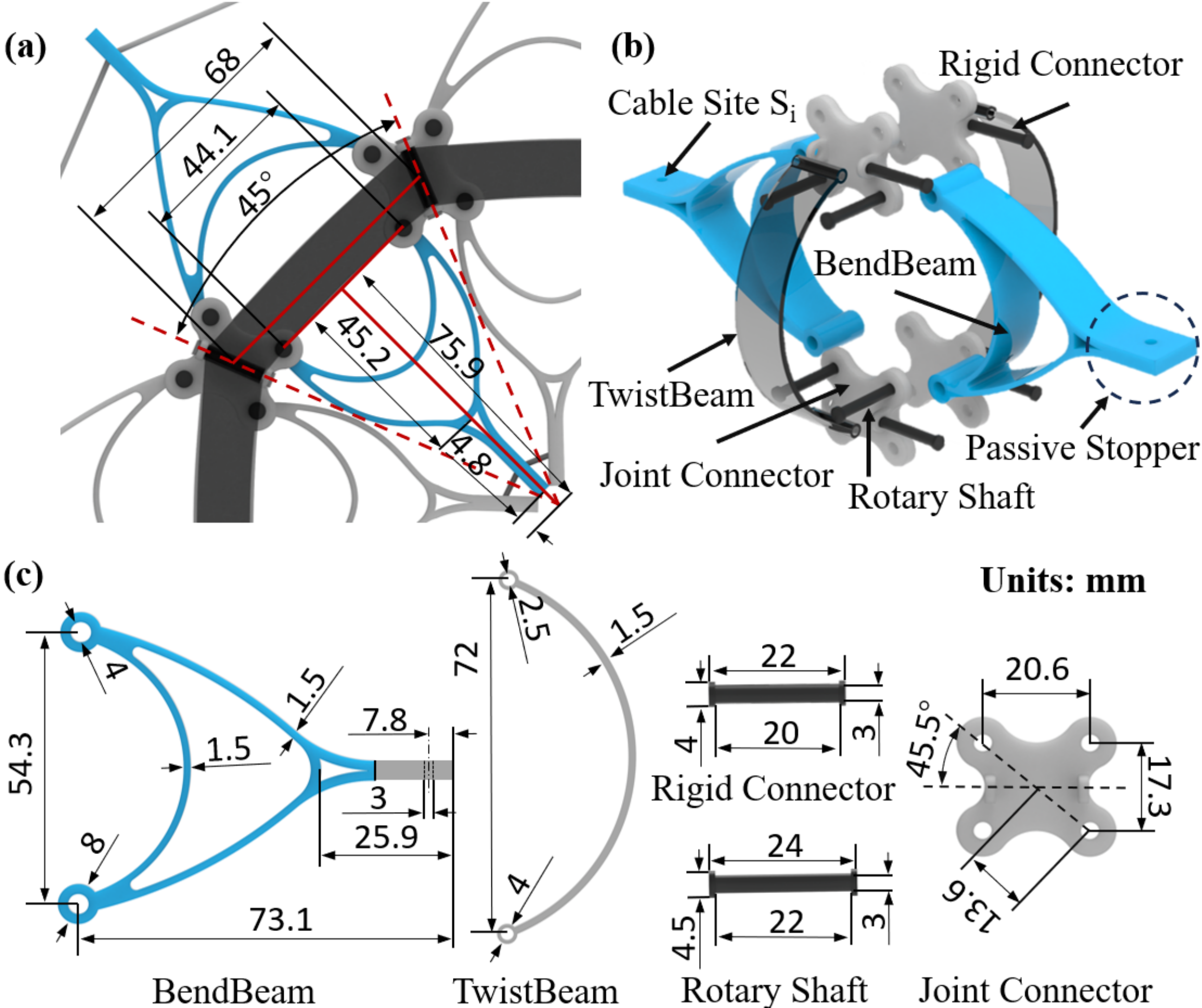} 
    \caption{(a) Geometric derivation of the passive stopper at the $45^\circ$ limit pose used in design; 
    (b) Exploded module showing BendBeam, TwistBeam, Joint Connector, Rigid Connector, and Rotary Shaft; 
    (c) Key dimensions of each component used to realize the passive limit.}
    \label{fig:module_geometry}
\end{figure}

Guided by the TwistBeam safety bound from combined bending–compression FEA ($45^\circ$ at $4\,\mathrm{mm}$ axial shortening), we designed a passive stopper mechanism to constrain module motion within a safe range. Each BendBeam carries a small stopper at its end; when the module approaches the limit pose, adjacent stoppers contact and halt further deformation.  

As shown in Fig.~\ref{fig:module_geometry}(a), at the $45^\circ$ limit the center-to-center distance between Rotary Shafts decreases from $54.3$ to $44.1\,\mathrm{mm}$, corresponding to $\Delta = 10.2\,\mathrm{mm}$ compression, well below the $25\,\mathrm{mm}$ fatigue-safe threshold validated in BendBeam tests. In this state, the flexible arc length is $L_f = 45.2\,\mathrm{mm}$ and the minimum radial clearance is $c = 4.8\,\mathrm{mm}$. With a total radial span $L_t = 75.9\,\mathrm{mm}$, the stopper length is designed as  
\[
L_s = L_t - L_f - c = 25.9\,\mathrm{mm}.
\]
This ensures stopper contact occurs before the TwistBeam exceeds its fatigue-safe range, providing passive fatigue-aware protection. Figures~\ref{fig:module_geometry}(b)–(c) illustrate the exploded view and key dimensions of the components.  

Beyond geometric protection, the stopper introduces a repeatable contact event exploitable for sensing. At the limit pose, displacement ceases while actuation torque rises sharply. Using motor-side torque sensing, this transition can be detected reliably without additional sensors. Thus, the passive stopper both constrains the operational boundary and provides a consistent physical reference for limit detection, enabling reliable force data acquisition for further real-time fatigue estimation.   


\section{Fatigue Awareness via Model-Based Estimation}\label{Op}
\begin{figure}[t]
  \centering
  \includegraphics[width=.7\linewidth]{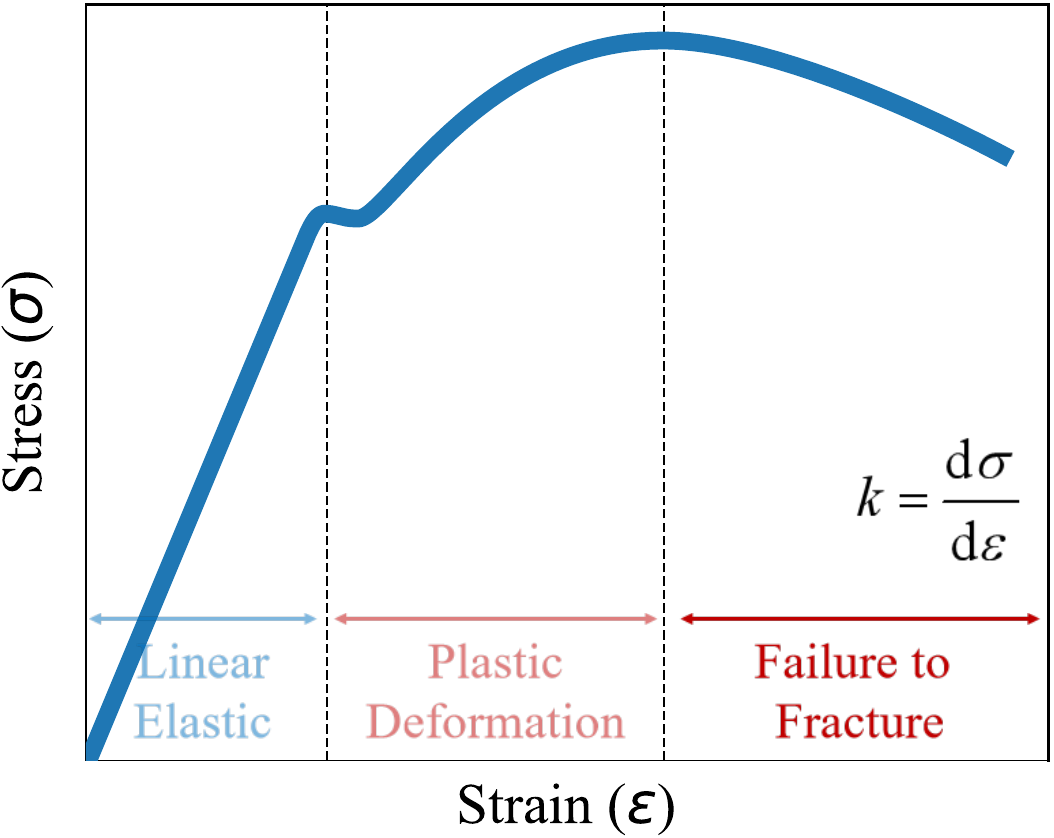}
  \caption{Stress-Strain Curve of Elastic Bodies}
  \label{fig:stress_strain}
\end{figure}

According to materials mechanics~\cite{callister2014}, the stiffness of an elastic material $k=\frac{\mathrm{d}\sigma}{\mathrm{d}\epsilon}$ remains constant in the linear elastic phase, decreases in the plastic phase, and becomes negative once damage occurs (Fig.~\ref{fig:stress_strain}). Since continuum robots essentially behave as elastic bodies, stiffness variation can in principle be exploited to estimate fatigue as well. By employing dynamic modeling in Multi-Joint dynamics with Contact (MuJoCo)~\cite{todorov2012mujoco,mujoco2021}, we enable model-based stiffness estimation to monitor fatigue progression.

\subsection{Dynamic Modeling}

The continuum robot consists of seven modules comprising BendBeams and TwistBeams (Fig. \ref{fig:hybrid_structure}). Following the pseudo rigid body model (PRBM) \cite{howell1996loop}, each BendBeam and TwistBeam can be represented by a limited number of rigid prismatic and revolute joints with passive stiffness, which is instinctively compatible with MuJoCo. 

\subsubsection{Model Formulation}
Applying the Newton–Euler formulation to each rigid body and joints yields the following nonlinear dynamics for the continuum robot (see details in \cite{todorov2014mujoco} and \cite{renda2014discrete}):
\begin{equation}
\small
M(q, \ddot{q})  + C(q, \dot{q}) + G(q) + K(q) + D(q, \dot{q}) = H(q)_c^{\mathrm{T}}u_c + H(q)_e^{\mathrm{T}}u_e
\end{equation}
where 
$q = \begin{bmatrix} \theta^\top & L^\top \end{bmatrix}^\top$, 
$q,\dot{q},\ddot{q} \in \mathbb{R}^{77}$ stand for the displacement, velocity and acceleration of the joints, 
with $\theta \in \mathbb{R}^{42}$ and $L \in \mathbb{R}^{35}$ respectively collecting all rotational and translational DOFs; 
$u_c \in \mathbb{R}^{2}$ is the control input and $u_e \in \mathbb{R}^{m_e}$ is the contact force here in this case including the self contact, 
with $H_c(q) \in \mathbb{R}^{2 \times 77}$ and $H_e(q) \in \mathbb{R}^{m_e \times 77}$ mapping to generalized forces; 
$M(q)$, $C(q,\dot{q})$, $K(q)$, $D(q,\dot{q}) \in \mathbb{R}^{77 \times 77}$, 
$G(q) \in \mathbb{R}^{77 \times 1}$ are the inertia matrix, Coriolis/centrifugal matrix, joint stiffness matrix, damping matrix, and gravitational force vector, respectively.

\begin{figure}[t]
    \centering
    \includegraphics[width=\columnwidth]{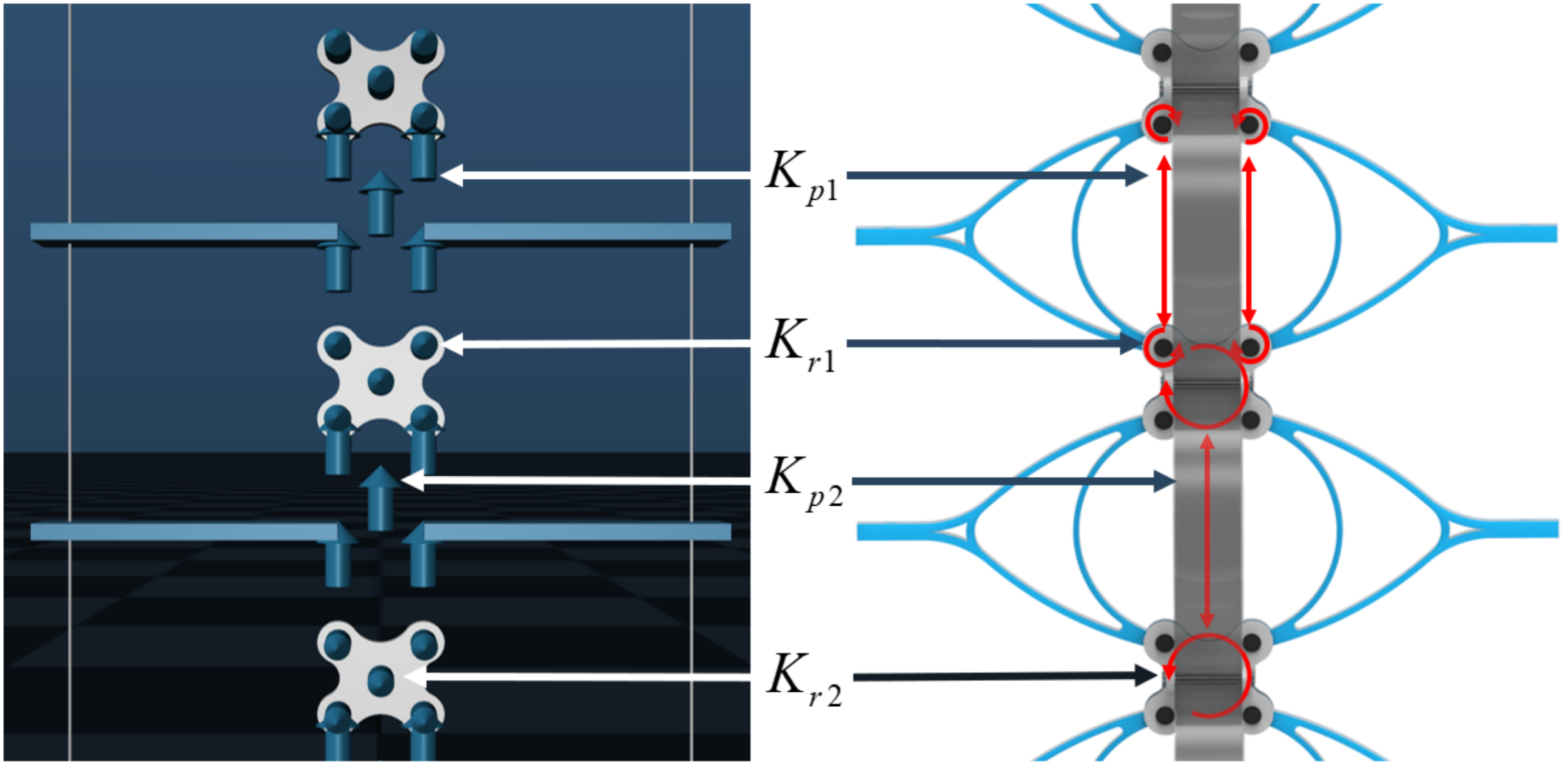} 
   \caption{Equivalent joint model of BendBeams and TwistBeams with stiffness parameters.}
    \label{fig:K}
\end{figure}

As illustrated in Fig.~\ref{fig:K}, each BendBeam is modeled as two compliant prismatic joints with translational stiffness \(K_{p1}\) connected to two passive revolute joints with \(K_{r1}=0\). The two TwistBeams per module are lumped into one compliant prismatic joint with stiffness \(K_{p2}\) and two passive revolute joints with rotational stiffness \(K_{r2}\). $K_{p1}$, $K_{p2}$, and $K_{r2}$ exhibit quadratic nonlinearity. According to the Stone–Weierstrass Theorem~\cite{rudin1991functional}, they can be approximated as $K_{p1} \approx \sum_{i=0}^{n} a_i L^i$, $K_{p2} \approx \sum_{i=0}^{n} b_i L^i$, and $K_{r2} \approx \sum_{i=0}^{n} c_i \theta^i$. The stiffness matrix $K(q)$ is defined as
\begin{equation}
\small
K(q)=f\!\left(q, K_{p1}, K_{p2}, K_{r2}\right) \in \mathbb{R}^{77\times77}.
\end{equation}
The parameters $K_{p1}$, $K_{p2}$, and $K_{r2}$ within the stiffness matrix are the key quantities to be identified, serving as indicators of the robot’s fatigue state.

\subsection{Fatigue Awareness}
To monitor fatigue, the robot’s stiffness parameters must be identified. We formulate stiffness identification as an optimization problem. A control input $u_c$ is applied to the real robot, and $u_c^*$ is recorded when it reaches the limit position ($q = q^*$), thereby perceiving the robot state. The same $u_c^*$ is then applied to the simulated robot, and $K(q^*)$ at the limit position is estimated by solving:
\begin{equation}
\small
\begin{aligned}
&(K_{p1}, K_{p2}, K_{r2}) = \arg\min \quad g(q^*)^2 \\
\text{s.t.} \quad G&(q^*) + K(q^*)q^* = H(q^*)_c^{\mathrm{T}}u_c^* + H(q^*)_e^{\mathrm{T}}u_e ,
\end{aligned}
\end{equation}
where $g(q^*) = \sum_{i=1}^{6}\left\|S_{i+1} - S_i\right\|$ is the sum of distances between cable sites $S_i$ on the BendBeams (Fig.~\ref{fig:module_geometry}b). The constraint corresponds to (2) under $q = q^*$, $\dot{q}=0$, $\ddot{q}=0$, and $u_c = u_c^*$, which physically implies that the robot is statically at its limit position (Fig.~\ref{fig:hybrid_structure}a). Again, a reduction in stiffness reflects fatigue accumulation, whereas a sharp drop indicates fracture (Fig.~\ref{fig:stress_strain}). Thus, stiffness serves as a key indicator for detecting the fatigue state of the robot.

\begin{figure}[t]
    \centering
    \includegraphics[width=\linewidth]{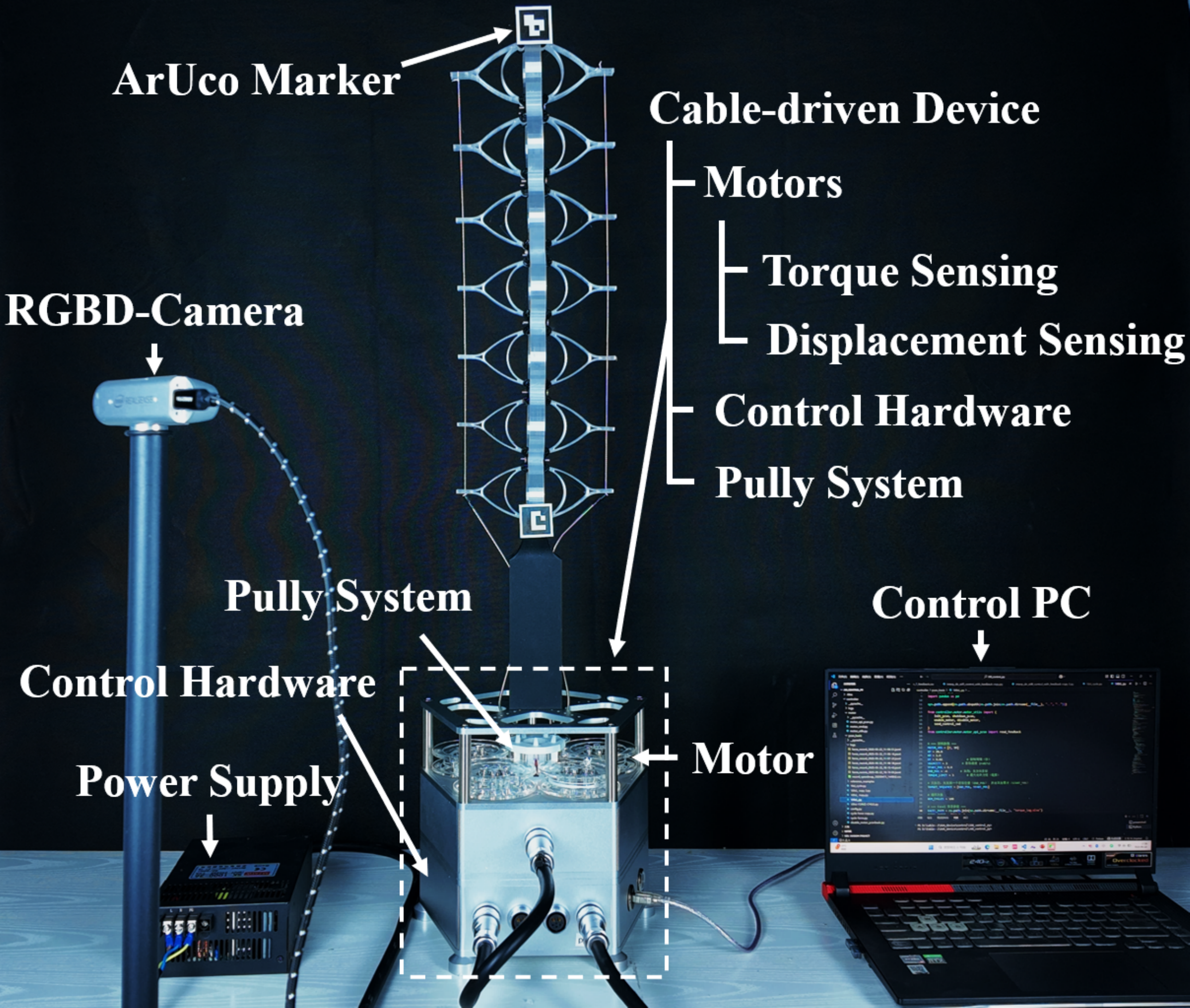}
    \caption{
    Experimental platform.
    }
    \label{fig:fig_cycle_force}
\end{figure}

\begin{figure*}[!h]
    \centering
    \includegraphics[width=0.95\linewidth]{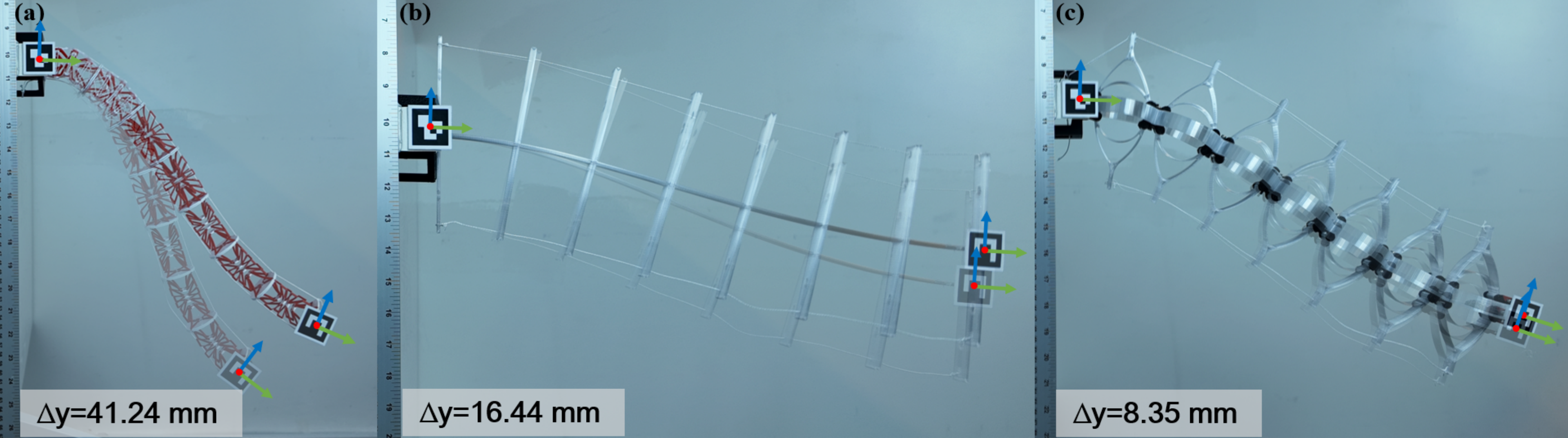}
    \caption{End-effector deflection before (solid) and after (transparent) 3,000 fatigue cycles for three robot designs.}

    \label{fig:rotational_hinge}
\end{figure*}


\section{Experimental Validation}
\label{sec:platform}
In this section, we validate the above proposed design and fatigue estimation strategy, including the Hybrid Hinge–Beam design for stress redistribution, the Passive Stopper for safe actuation and stable feedback, and the model-based fatigue awareness strategy. The experimental platform is shown in Fig.~\ref{fig:fig_cycle_force} and consists of two modules:  
(i) a cable-driven device integrating motors, control hardware, and a pulley system, which enables torque sensing and displacement sensing from built-in encoders; and  
(ii) a vision-based monitoring module employing ArUco markers and a RGBD Camera ~\cite{garrido2014automatic} for real-time pose and trajectory tracking.  



\subsection{Fatigue-Induced Drift Under Cyclic Loading}

This fatigue test evaluates the impact of different backbone topologies, including prior designs~\cite{dewi2024modular,walker2013continuum} and the proposed design, on fatigue performance. Three continuum robots were constructed for comparison, each representing a distinct structural philosophy. 

\subsubsection{Experimental Setup and Procedure}
\begin{itemize}
    \item \textbf{Reference design}~\cite{dewi2024modular}: Incorporates multiple thin flexure hinges within each modular joint to redistribute strain and mitigate stress concentration, thereby enhancing fatigue performance.  
    \item \textbf{Conventional design}~\cite{walker2013continuum}: A classical single-backbone configuration widely adopted in continuum robotics, serving here as a baseline for comparison.  
    \item \textbf{Proposed design}: A hybrid hinge–beam topology that integrates passive revolute joints and TwistBeams to redistribute stress, aiming to improve fatigue resistance while preserving stiffness.  
\end{itemize}
\begin{table}[t]
\centering
\caption{NTDR After Fatigue Cycles}
\label{tab:ntdr_results}
\begin{tabular}{|c|c|c|c|}
\hline
Cycle Number & Reference &Conventional & Proposed \\
\hline
1,000 & 0.0142 & 0.0085 & 0.0051 \\
2,000 & 0.0438 & 0.0191 & 0.0109 \\
3,000 & 0.0957 & 0.0365 & 0.0185 \\
\hline
\end{tabular}
\end{table}

All three designs were fabricated from PETG through 3D printing, with total lengths of 430.9~mm (Reference) and 450.4~mm (Conventional and Proposed). Each prototype was driven with identical cyclic inputs: from neutral to 180° left bending, then to 180° right, and back to neutral. Tests were conducted after 1k, 2k, and 3k cycles. After each session, tendon lengths were reset and the robot was fixed horizontally.


Fatigue-induced drift was evaluated by measuring the difference between the final and initial y-axis tip offsets, denoted as $\Delta y$. In addition, the Normalized Tip Deflection Ratio (NTDR) was defined as
\begin{equation}\label{eq:ntdr}
    \mathrm{NTDR} = \frac{|\Delta y|}{L},
\end{equation}
where $L$ is the total length of each robot, providing a normalized basis for comparison. To ensure consistency, the initial displacement input was kept constant, and $\Delta y$ was measured as the difference between the first and last y-axis tip positions under continuous tendon loading.

\subsubsection{Results and Discussion}
Results are summarized in Table~\ref{tab:ntdr_results} and Fig.~\ref{fig:rotational_hinge}. After 3,000 cycles, the reference, conventional, and proposed designs yielded $\Delta y = 41.24$\,mm, $16.44$\,mm, and $8.35$\,mm, respectively, representing a 79.8\% reduction relative to the reference and 49.2\% relative to the conventional design. Across 1k, 2k, and 3k checkpoints, the proposed design consistently achieved the lowest NTDR values (0.0051, 0.0109, and 0.0185), compared with 0.0365 (conventional) and 0.0957 (reference) at 3k. At 3k cycles, the NTDR of the reference was 5.17$\times$ that of the proposed (80.7\% reduction), and the conventional was 1.97$\times$ (49.3\% reduction). The reduced drift and consistently lower NTDR validate the stress-redistribution intent of the hybrid hinge–beam structure: passive revolute joints alleviate bending-induced stress concentration in BendBeams, while the TwistBeam shares axial load and stabilizes orientation, delaying fatigue-driven permanent offsets. The monotonic trends across all cycles further support the robustness of these findings and align with the visual overlays in Fig.~\ref{fig:rotational_hinge}.

\subsection{Passive Stopper Limit Detection via Torque Sensing}  
The passive stopper acts as a geometric motion limiter combined with motor torque sensing. It not only constrains the actuation space for safe operation but also provides reliable feedback at the limit pose. This experiment validates the stopper’s ability to ensure safety and to generate distinct sensing signatures that indicate engagement of the mechanical limit.  

\subsubsection{Experimental Setup and Procedure}  
Motor-side torque sensing was employed to record cable displacement and payload torque, eliminating the need for external sensors. Velocity control was used to drive the robot, and both displacement and torque were logged at 30~Hz. The robot was actuated until the passive stopper engaged, ensuring that the post-contact torque response was captured. Each trial was repeated ten times to confirm repeatability. During this process, cable displacement was monitored to identify plateau events indicating motion cessation, while motor torque was simultaneously tracked to detect the sharp slope increase at the instant of contact (Fig.~\ref{fig:constraint_validation}). The sampling rate of 30~Hz was sufficient to capture these transitions in real time.  

\subsubsection{Results and Discussion}  
\begin{figure}[t]
    \centering
    \includegraphics[width=\columnwidth]{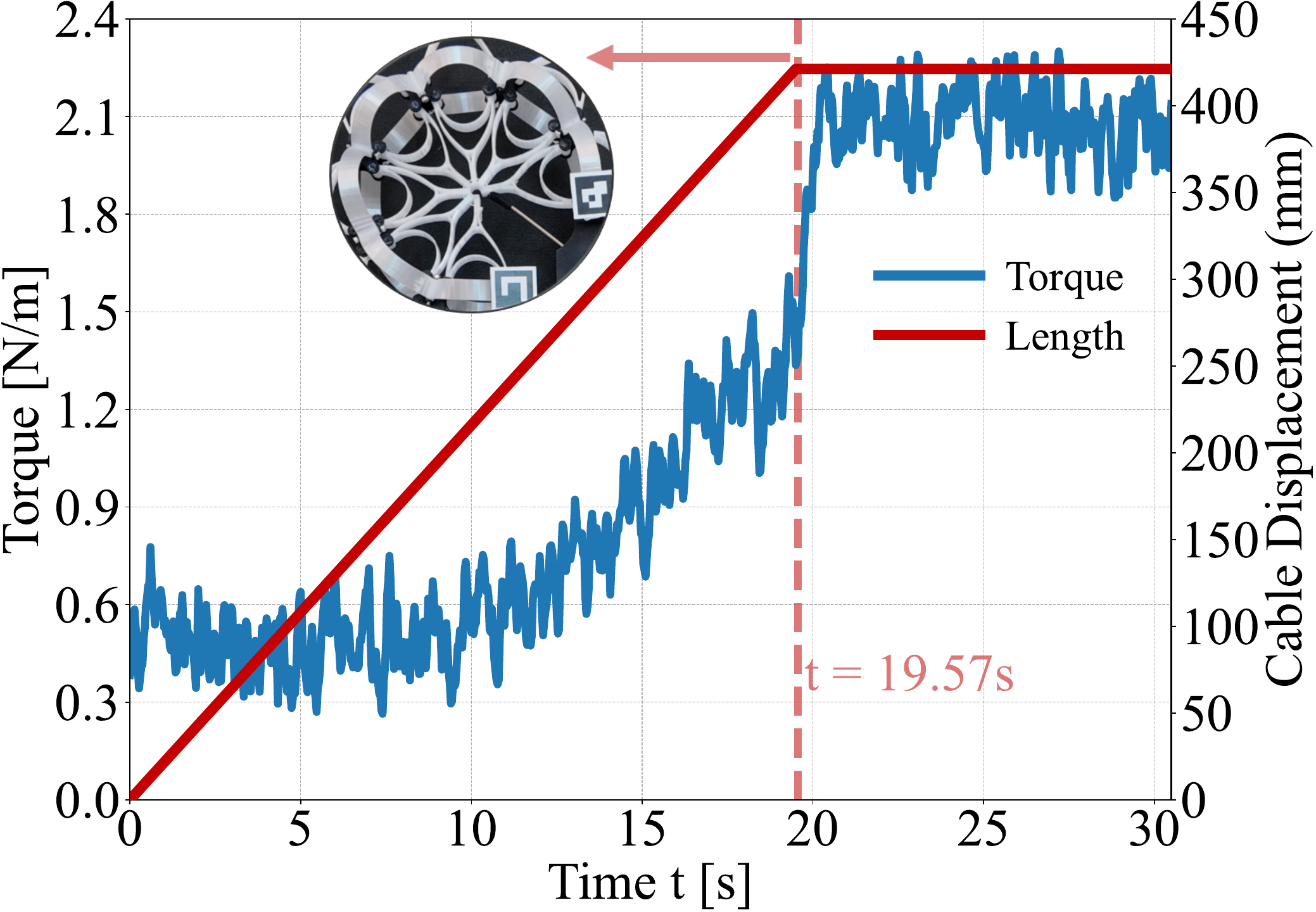}
    \caption{Validation of the passive geometric constraint: torque and cable displacement versus time $t$. }
    \label{fig:constraint_validation}
    \vspace{-4mm}
\end{figure}

The results are shown in Fig.~\ref{fig:constraint_validation}. A distinct state transition occurs when the stopper engages at $t \approx 19.57$~s:  
\begin{itemize}  
    \item \textbf{Displacement Response:} Cable displacement (red, right axis) increases linearly during actuation. At $t \approx 19.57$~s, it abruptly plateaus at 422~mm, indicating cessation of motion at the mechanical limit.  
    \item \textbf{Torque Response:} Motor torque (blue, left axis) increases gradually during free motion. At contact ($t \approx 19.57$~s), the slope rises sharply, reaching 1.4~$\mathrm{N\cdot m}$. This clear transition provides a robust trigger metric for detecting engagement of the mechanical limit.  
\end{itemize}

Fig.~\ref{fig:constraint_validation} provides a clear visual confirmation of the robot’s state at contact. The photograph shows the passive stoppers fully engaged, verifying that the sharp transition in the data corresponds directly to this mechanical event. The results were highly repeatable across all trials, with the trigger torque consistently measured at $-1.4 \pm 0.02\,\mathrm{N\!\cdot\! m}$ (mean $\pm$ std. dev.). These findings validate that the geometric constraint serves as a robust and precisely detectable physical marker for the perception algorithm, correlating a distinct data signature with a confirmed mechanical event while defining the robot’s safe operational domain.

\subsection{Validation of Fatigue Detection Feasibility}
\label{subsec:fatigue_self_sensing}

To investigate the fatigue process of the continuum robot, we conducted cyclic fatigue-to-failure experiments and, based on the data, established the relationship between the limit pose torque $\tau_{\mathrm{lim}}$ and the estimated stiffness $K(q^*)$ (obtained via the model-based estimation in Section~\ref{Op}). This enables real-time fatigue detection by directly inferring stiffness from $\tau_{\mathrm{lim}}$.  

\subsubsection{Experimental Setup and Procedure}  
In each cycle, the robot was driven from rest to the mechanical limit pose defined by the passive stopper (corresponding to a 270$^\circ$ tip rotation), and then back to rest. During each cycle, the motor torque profile was recorded. The torque value at the limit pose was identified as $\tau_{\mathrm{lim}}$ and stored for every cycle. In total, more than $n=9{,}000$ cycles were performed.  
\begin{figure}[t]
    \centering
    \includegraphics[width=0.90\linewidth]{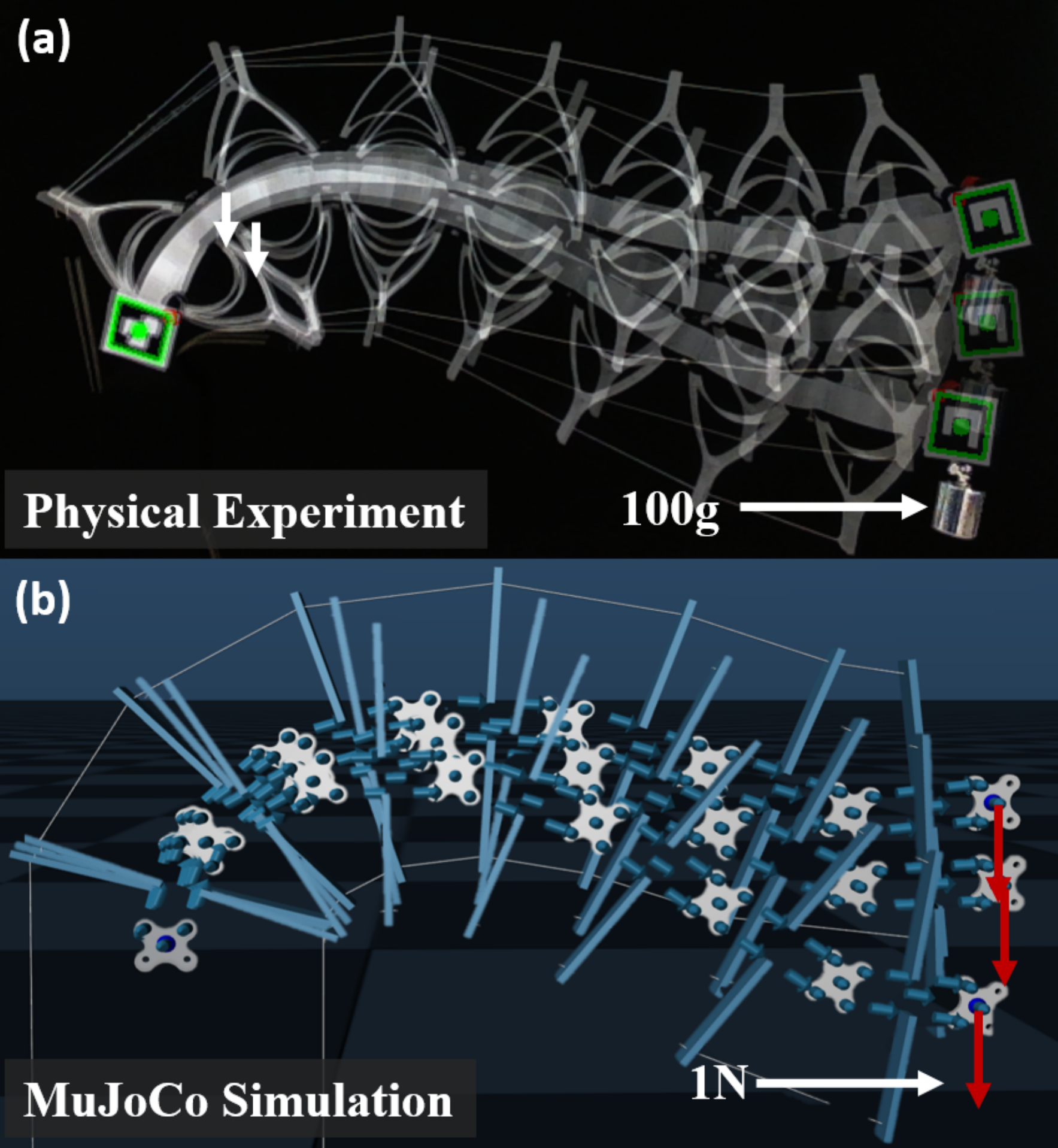}
    \caption{
    Experimental platform.
    (a) physical experiment with Aruco marker tracking used to obtain position information;
    (b) corresponding MuJoCo simulation scene.
    }
    \label{fig:fig_force_mujoco1}
\end{figure}
Before establishing the relationship between $\tau_{\mathrm{lim}}$ and $K(q^*)$, stiffness estimation verification was performed at $n=3{,}000$, $6{,}000$, and $9{,}000$ cycles. In each case, the robot was actuated with the same cable displacement input, and a 100\,g payload was attached at the tip (equivalent to a 1\,N downward force in the model) to test whether the estimated stiffness could still reflect the true state under external loading. As shown in Fig.~\ref{fig:fig_force_mujoco1}, the results confirmed the effectiveness of the estimation strategy, before which fatigue estimation experiments were conducted.   

\begin{figure}[t]
    \centering
    \includegraphics[width=\linewidth]{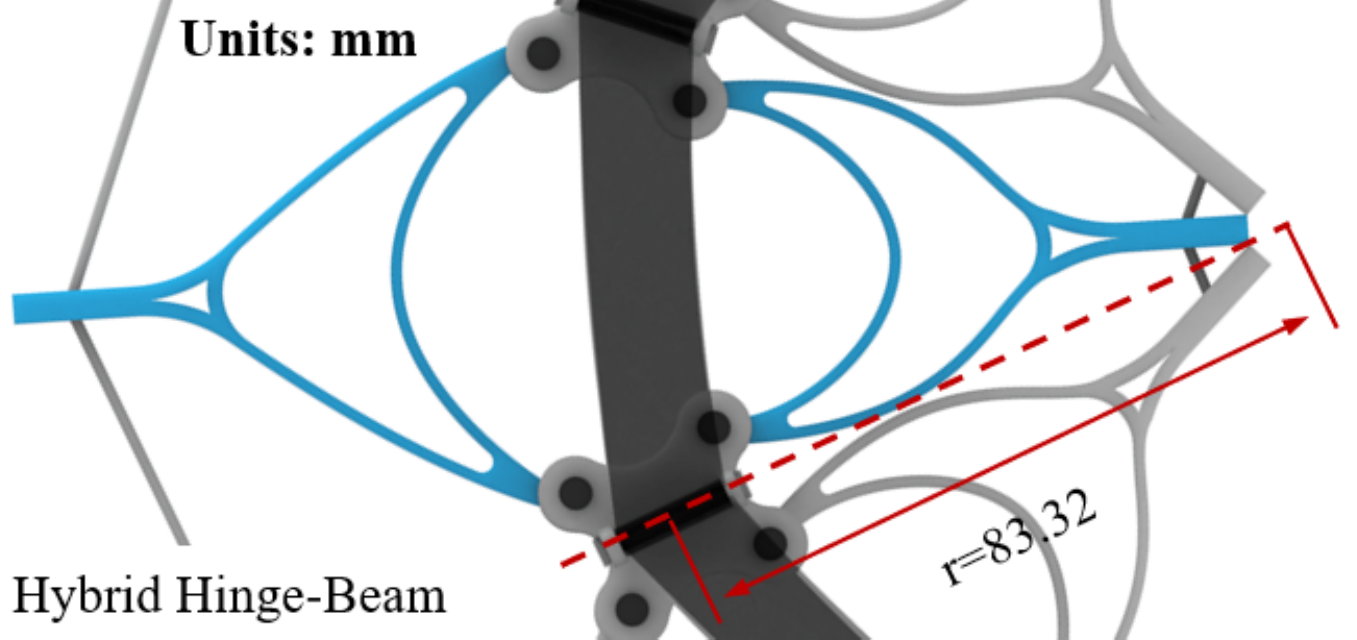}
    \caption{
Characteristic Length  of Hybrid Hinge-Beam
    }
    \label{fig:fig_force_mujoco}
\end{figure}

The primary metric is the limit pose torque $\tau_{\mathrm{lim}}(n)$. For stiffness estimation, a feature point was taken every 500 cycles, and offline model-based estimation of $K(q^*)$ was performed at the known geometric limit pose $q^*$. This yields an equivalent scalar stiffness $\hat K(n)$, defined as  
\begin{equation}
\small 
    \hat K = \sqrt{\frac{K_{p1}^2 + K_{p2}^2 + K_{r2}'^2}{3}},
\end{equation}
where $K_{r2}' = \frac{K_{r2}}{r^2}$ normalizes the units to N/m, ensuring $K_{p1}$, $K_{p2}$, and $K_{r2}$ are comparable in scale. The reference radius $r$, representing a characteristic length of the Hybrid Hinge-Beam structure, is selected as shown in Fig.~\ref{fig:fig_force_mujoco}. 
\begin{figure}[t]
    \centering
    \includegraphics[width=\linewidth]{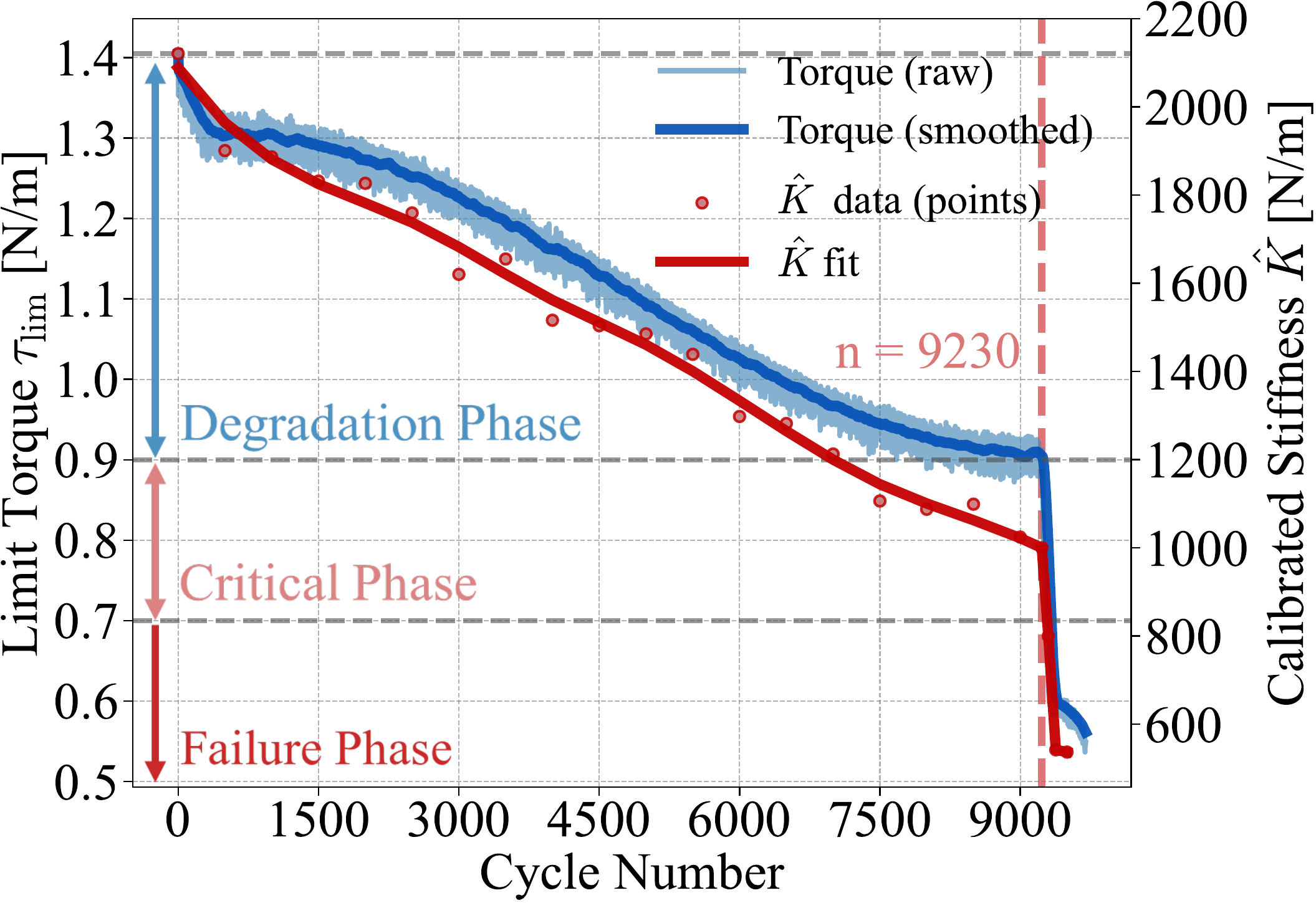}
   \caption{Dual-axis plots of the calibrated stiffness $\hat K(n)$ and limit torque $\tau_{\mathrm{lim}}(n)$ with phase divisions.}

    \label{fig:fatigue_dualaxis}
\end{figure}

\subsubsection{Results and Discussion}
Because real-time stiffness estimation is computationally expensive, we fitted a mapping $\hat K \approx f(\tau_{\mathrm{lim}})$ from the 500-cycle feature points to enable runtime estimation. Fig.~\ref{fig:fatigue_dualaxis} shows the dual-axis plots of $\tau_{\mathrm{lim}}(n)$ (blue, left axis) and $\hat K(n)$ (red, right axis). To improve curve accuracy, additional sampling points were introduced in regions with rapid variation, supplementing the baseline of one sample every 500 cycles. To examine correlation, both $\tau_{\mathrm{lim}}(n)$ and $\hat K(n)$ were smoothed and their second derivatives $\mathrm{d}^2/\mathrm{d}n^2$ computed. The results reveal highly consistent trends across the entire lifespan, with a synchronous sharp change at $n=9230$, coinciding with the onset of observable TwistBeam fatigue. Beyond this point, both metrics enter a rapid degradation phase until fracture, after which $\tau_{\mathrm{lim}}$ stabilizes at $\sim 0.6\,\mathrm{N\cdot m}$. These results show that $\tau_{\mathrm{lim}}(n)$ and $\hat K(n)$ follow identical monotonic trends, enabling limit torque to serve as a practical surrogate for stiffness in real-time fatigue estimation. Thus, the offline stiffness-based approach can be effectively replaced by $\tau_{\mathrm{lim}}$ for runtime perception.  
\begin{figure}[t]
    \centering
    \includegraphics[width=\columnwidth]{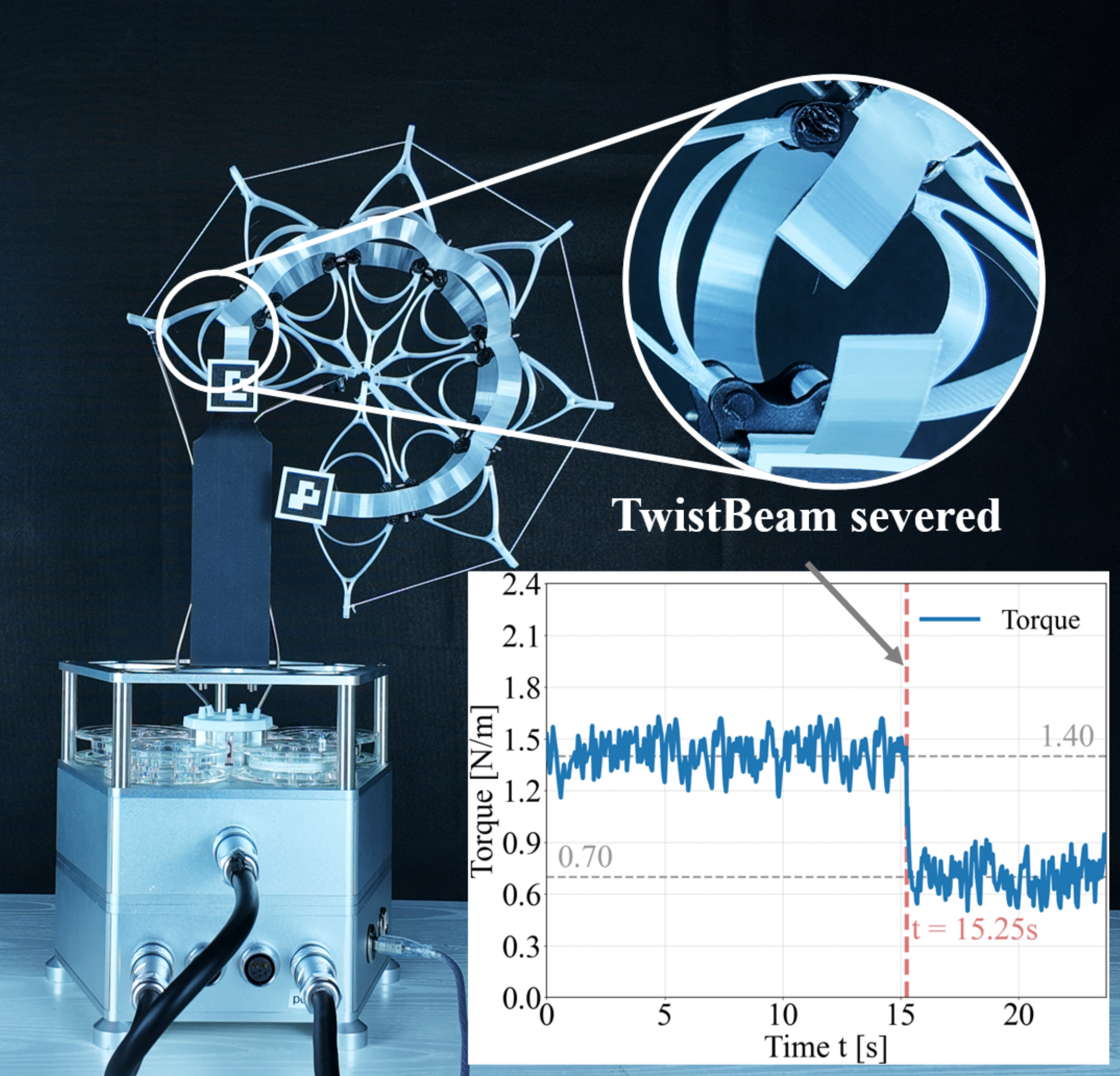}
    \caption{Measured torque response before and after TwistBeam severing. }
    \label{fig:FIG-17}
\end{figure}
To define a clearer boundary for the \textit{Fracture Stage}, a new robot was driven to its limit pose and the TwistBeam was manually severed, yielding $\tau_{\mathrm{lim}} \!\approx\! 0.7\,\mathrm{N\cdot m}$ as shown in Fig.~\ref{fig:FIG-17}. Accordingly, three regimes are delineated:  
\begin{itemize}
    \item \textbf{Degradation Phase} ($0.9 \leq \tau_{\mathrm{lim}} \leq 1.4$ $\mathrm{N\cdot m}$: Gradual stiffness loss prior to critical damage. The lower bound ($0.9$ $\mathrm{N\cdot m}$) corresponds to a sharp trend change, similar to that observed at $N=9230$ in Fig.~\ref{fig:fatigue_dualaxis}.  
    \item \textbf{Critical Phase} ($0.7 \leq \tau_{\mathrm{lim}} < 0.9$ $\mathrm{N\cdot m}$): Rapid stiffness degradation with severe fatigue symptoms.  
    \item \textbf{Failure Phase} ($\tau_{\mathrm{lim}} < 0.7$ $\mathrm{N\cdot m}$): Imminent fracture with negligible remaining load capacity.  
\end{itemize}

By delineating the fatigue process into distinct phases and leveraging $\tau_{\mathrm{lim}}$ as a runtime perception metric, real-time fatigue estimation becomes feasible. This allows the robot’s operational state to be continuously assessed and provides a quantitative basis for determining whether the system remains within a safe regime for continued operation. In practice, such online monitoring enables predictive maintenance, timely intervention before catastrophic failure, and more reliable long-term deployment of continuum robots in safety-critical tasks.



\section{Conclusions and Future Work}

This paper presented a fatigue-aware cable-driven continuum robot (CDCR) that integrates three tightly coupled advances. First, a Hybrid Hinge–Beam structure decouples torsion and bending: BendBeams with passive revolute joints relieve bending-induced stress concentration, while TwistBeams provide controlled torsional compliance and axial load sharing, further promoting uniform stress distribution. Second, a Passive Stopper imposes a geometric motion limit and yields a repeatable contact event whose motor-side torque signature enables precise limit detection and safe workspace enforcement. Third, a model-based fatigue-awareness method estimates stiffness from the limit-pose torque, allowing online, sensor-free monitoring of fatigue progression. FEA and bench-top experiments corroborate the design, showing a \mbox{49\%} reduction in fatigue accumulation relative to a conventional baseline, together with real-time model-based fatigue estimation. Future work will extend the framework to fully three-dimensional, multi-section CDCRs and investigate lightweight learning modules to refine stiffness estimation under varying loads and environments.

\bibliographystyle{IEEEtran}
\bibliography{refs}

\end{document}